\documentclass[sn-mathphys,Numbered]{sn-jnl}
\usepackage[utf8]{inputenc} 
\usepackage{mdframed}

\usepackage{hyperref}
\hypersetup{
    colorlinks=true,
    linkcolor=blue,
    filecolor=magenta,      
    urlcolor=cyan,
    pdftitle={Causality4Agriculture Perspective},
    pdfpagemode=FullScreen,
    }
\usepackage{comment}
\usepackage{graphicx}%
\usepackage{multirow}%
\usepackage{amsmath,amssymb,amsfonts}%
\usepackage{amsthm}%
\usepackage{mathrsfs}%
\usepackage[title]{appendix}%
\usepackage{xcolor}%
\usepackage{textcomp}%
\usepackage{manyfoot}%
\usepackage{booktabs}%
\usepackage{algorithm}%
\usepackage{algorithmicx}%
\usepackage{algpseudocode}%
\usepackage{listings}%
\usepackage{longtable}
\usepackage{multirow}
\usepackage{colortbl} 

\usepackage{xcolor}
\definecolor{nice-green}{HTML}{007849}
\definecolor{nice-blue}{HTML}{0375B4}
\definecolor{nice-orange}{HTML}{CC7722}
\definecolor{nice-red}{HTML}{FF5733}
\makeatletter
\newcommand{\customlabel}[2]{%
   \protected@write \@auxout {}{\string \newlabel {#1}{{#2}{\thepage}{#2}{#1}{}} }%
   \hypertarget{#1}{}
}
\makeatother

\usepackage{verbatim}

\begin{document}

\UseRawInputEncoding
\title[Causal ML for Agriculture and Food]{Causal Machine Learning for Sustainable Agroecosystems}


\author*[1]{Vasileios Sitokonstantinou} \email{Vasileios.Sitokonstantinou@uv.es}
\author[1]{Emiliano D\'iaz Salas-Porras}
\author[1]{Jordi Cerd\`a-Bautista}
\author[1]{Maria Piles}
\author[2]{Ioannis Athanasiadis}
\author[4]{Hannah Kerner}
\author[3]{Giulia Martini}
\author[5]{Lily-belle Sweet}
\author[2,6]{Ilias Tsoumas}
\author[5]{Jakob Zscheischler}
\author[1]{Gustau Camps-Valls} 

\affil*[1]{\orgdiv{Image Processing Laboratory}, \orgname{Universitat de Val\`encia}, 
\country{Spain}}

\affil[2]{Artificial Intelligence, Wageningen University and Research, The Netherlands}
\affil[3]{World Food Program, UN, Rome, Italy}
\affil[4]{Arizona State University, USA}
\affil[5]{Department of Compound Environmental Risks, Helmholtz Centre for Environmental Research -- UFZ, Leipzig, Germany}
\affil[6]{National Observatory of Athens, Greece}

\abstract{Sustainable agriculture is essential for food security and environmental health in a changing climate. However, it is challenging to understand the complex interactions among its biophysical, social, and economic components. Predictive machine learning (ML), with its capacity to learn from data, is leveraged in sustainable agriculture for applications like yield prediction and weather forecasting. Nevertheless, it cannot explain causal mechanisms and remains descriptive rather than prescriptive. To address this gap, we propose causal ML, which merges ML's data processing with causality's ability to reason about change. This facilitates quantifying intervention impacts for evidence-based decision-making and enhances predictive model robustness. We showcase causal ML through eight diverse applications that benefit stakeholders across the agri-food chain, including farmers, policymakers, and researchers.}

\maketitle

\section{Introduction}

The perception of agriculture is evolving. Nowadays, policymakers, researchers, farmers, and consumers recognize farms as integral components of larger, interconnected agroecosystems that include biological, physical, social, and economic dimensions \cite{francis2003agroecology}. However, current agricultural practices often fail to balance these elements, relying heavily on agricultural land expansion, external inputs like synthetic fertilizers and pesticides, and resource extraction. These practices contribute to environmental impacts like biodiversity loss and soil degradation that pose significant threats to future generations \cite{pretty2008agricultural}.

Sustainably achieving food security targets is a complex task. Production must be intensified to meet the growing food demand \cite{godfray2010food, van2021meta}, even on lands already degraded \cite{eswaran2019land}, while simultaneously reducing greenhouse gases emissions \cite{ivanovich2023future}.
This challenge is compounded by slow technology adoption \cite{takahashi2020technology}, inefficient policies \cite{peer2020european}, and unforeseen crises like pest outbreaks, pandemics, and financial disruptions \cite{altieri2020agroecology}. Beyond achieving intensification without harming the environment and adding to climate change, agriculture has the potential to mitigate these problems by sequestering carbon in soils and enhancing biodiversity. Thus, agriculture can be both a contributor to and a solution for climate change \cite{umesha2018sustainable}. To harness agriculture's potential for positive impact, we need a comprehensive framework that evaluates strategies based on their contributions to sustainability across the entire agroecosystem. Improved understanding of sustainability across all dimensions of agriculture can help prioritize the most impactful actions \cite{gliessman2016transforming}.

In this context, where agricultural decisions have far-reaching impacts, traditional approaches to modeling and decision-making may struggle to capture the complexity of agroecosystems. Machine learning (ML) methods have emerged as powerful tools for finding patterns within large datasets and making predictions based on historical data \cite{meshram2021machine}. However, while ML excels at predicting outcomes, it cannot explain the underlying causality, which limits its effectiveness in performing robustly in new, changing environments \cite{scholkopf2022causality} and evaluating the impact of interventions (see Box \ref{box1}).
 
\begin{figure}[t!]
  \centering
  \includegraphics[width=1 \textwidth]{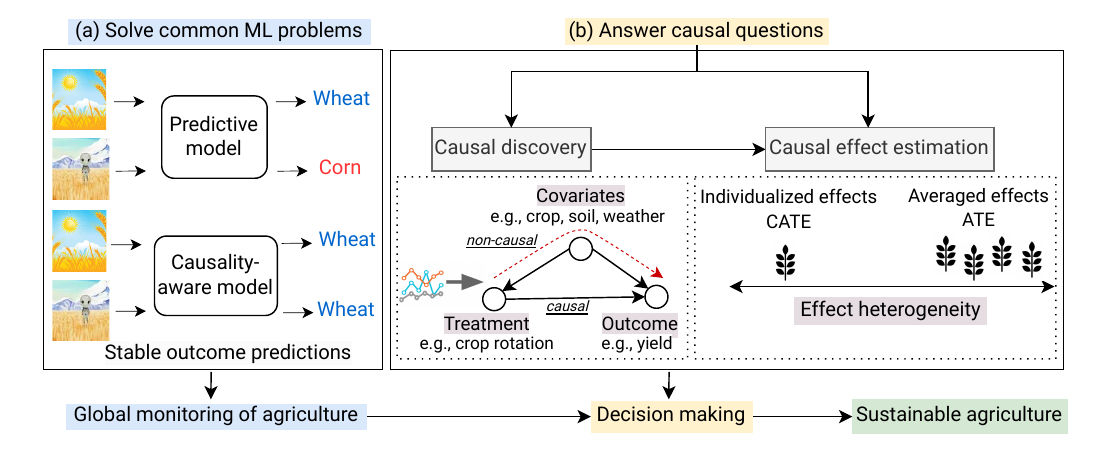}
  \caption{\textbf{How causality can drive sustainability in agriculture:} a) Solve common ML problems: Address robustness to interventions over time (e.g., new policies) and geographic generalization to develop predictive models that can continuously and globally predict agricultural activity and ecological conditions. For instance, a standard model might fail to recognize wheat in snowy conditions if it was trained only on sunny conditions. A causality-aware model, however, learns the stable underlying features of wheat, allowing accurate identification in any weather. b) Answer causal questions: Use large-scale, robust predictions from (a) and/or other agricultural observations with causal discovery methods to produce causal graphs that map cause-effect relationships. Based on these graphs, estimate the effects of treatments (averaged or individualized) on outcomes of interest, as the relevant covariates for isolating non-causal associations (red line) are now known. With this knowledge, the most effective solutions can be identified and prioritized.}
  
  \label{fig1}
\end{figure}

To address the limitations of predictive ML
and enhance our understanding of causal mechanisms in agricultural systems, we introduce causal ML (see Fig. \ref{fig1}). Causal ML includes methods designed to infer causality from data, but also leverage causal knowledge to enhance predictive ML models \cite{kaddour2022causal}. 
Despite its potential, the adoption of causal ML in agriculture has been slower compared to related disciplines like ecology \cite{runge2023modern}, public policy \cite{fougere2019causal}, and Earth and climate sciences \cite{runge2023causal}. This paper aims to bridge this gap by expressing complex ideas in a language that stakeholders in the food and agriculture sectors can understand and use. Integrating causal ML into decision-making processes can facilitate a transition towards evidence-based decisions that improve the sustainability of our food systems.

\clearpage 
\begin{mdframed}[backgroundcolor=gray!10]
\customlabel{box1}{1}
\textbf{Box 1: Key differences between predictive ML and causal ML}\\

\noindent{\bf Predictive ML} focuses on finding statistical associations or dependencies among input variables to make predictions of an outcome. 

\noindent{\bf Causal ML} is the synergistic use of causal inference and machine learning to either, improve causality with ML (ML for causality) or improve ML with causality (causality for ML).

\centerline{\includegraphics[width=4cm]{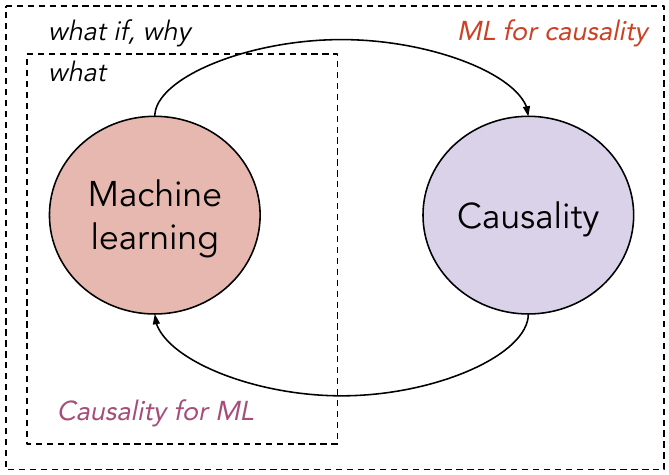}}

\noindent{\bf Causal questions:} Causal questions such as \emph{what happens if...} or \emph{what would have happened if...} can be expressed using Pearl's Structural Causal Model (SCM) framework \cite{pearl2009causality}. In this framework, a system is described with interdependent fundamental processes and random quantities. Each process is represented by an equation that includes qualitative elements, namely a list of causative variables, and quantitative elements, namely functions that describe relationships between the process variable and its causes. ML for causality involves two main sub-tasks: causal discovery, which determines SCM's qualitative elements, and causal effect estimation, which addresses SCM's quantitative elements.
\begin{itemize}
    \item {\em Question.} What is the effect of a crop rotation on the field's soil health?
    \item {\em Predictive ML} can detect correlations based on historical data and estimate soil health from crop rotation data. However, it may mistakenly attribute soil health changes to crop rotation.
    
    \item {\em ML for causality} allows us to express the decisions as causal questions and to use ML guided by causal reasoning to answer them,  i.e. to quantify to what extent the observed changes in soil health can be attributed to crop rotations.
\end{itemize}

 \noindent{\bf Predictive tasks:} Use the patterns in a training dataset to model associative relationships to predict outcomes in a test dataset. 

\begin{itemize}
    \item {\em Task.} Predict yield in Africa by exploiting the relationship between yield and biomass observed in Europe. 
    \item {\em Predictive ML.} It exploits all patterns produced by associative relationships to obtain accurate models. However, it relies on training and test data being very similar. It cannot discriminate between types of association (e.g., cause-effect) to generalize to new data distributions. 
    \item {\em Causality for ML.} Focuses on causal features with higher potential for generalization and robustness.
\end{itemize}

\noindent

\noindent 

\end{mdframed}

\section{Causal ML for agriculture}
\subsection{ML for answering causal questions}\label{sec:2}

Researchers have traditionally used randomized experiments to answer causal questions, namely establishing causal links between variables 
or estimating the effects of various practices or policies (treatments) 
\cite{petersen1994agricultural}. Examples of such questions are provided in  Box \ref{box2}. While these experiments are known for providing unbiased estimates, they come with important challenges, including small sample sizes that lead to less precise results \cite{kluger2022combining} and findings that might not apply universally to (even slightly) different conditions \cite{rothwell2005external}. Advancements in big data and satellite-based remote sensing now enable the investigation of the impacts of agricultural practices or policies on sustainability outcomes using extensive datasets that cover diverse conditions. Deep learning techniques, for example, allow for the efficient processing of large amounts of data from multiple sources at different spatiotemporal scales \cite{li2022deep}. These techniques enable the transformation and reduction of the raw data into useful representations. However, these learned representations are not necessarily aligned with the physical variables that best describe the system.
Another issue is that these datasets are observational, meaning they lack randomized treatments and rely on the assumption that no external factors are influencing both the treatment and the outcome. Violating these assumptions can lead to biased estimates of causal effects.

To exploit ML's potential while addressing these issues we introduce the following causal ML workflow for agriculture: 

\begin{itemize}
    \item {\bf Defining the Causal Question.} A causal question can be either qualitative or quantitative. For instance, a qualitative question might explore whether there is a causal relationship between soil microbiome diversity and crop yield. Such questions are addressed using causal discovery methods (see apps in Sec. \ref{discovery}). On the other hand, a quantitative question might seek to determine the extent of the impact, such as estimating how much the use of a particular pesticide increased the crop yield. These types of questions are tackled using causal effect estimation methods (see apps in Sec. \ref{policy} and \ref{farmers}). 
    
\item {\bf Collecting data.} To answer causal questions, we need to collect data on the agricultural system under study. 
To determine if cause-effect relationships exist between a group of variables (causal discovery), we need information on the variables of interest and any potential confounding variables: variables that may cause at least two of the variables of interest. To estimate the effectiveness of practices, policies, or technologies (causal effect estimation), we require data on the treatment of interest (e.g., fertilizer application), the observed outcome (e.g., farm profitability) for a specific unit (e.g., a field), and potential confounders like environmental factors (e.g., temperature), and crop and soil characteristics \cite{pearl2009causality}. This information can be sourced from remote sensing observations, reanalysis data, process-based model simulations, farm management systems, crop calendars, farm accounting, agricultural field experiments, and plant phenotypical platforms.

\item {\bf Making assumptions.} To answer causal questions without randomized experiments, we must make some assumptions, either structural, process, or statistical. 
{\em Structural assumptions} concern the nature of potential causal relationships: Is unmeasured confounding present? What is the maximum delay for cause-effect relationships? {\em Process assumptions}, for example, might question whether linear models are sufficient for capturing causal relationships.
{\em Statistical considerations} focus on evaluating whether data is adequate and representative of key variables. The core assumption for causal discovery is causal sufficiency (structural), ensuring all relevant variables are measured. For causal effect estimation, the key assumption is the causal graph (structural), which outlines all cause-effect relationships and can be derived from various sources including expert knowledge and data-driven methods. Rubin's potential outcomes framework \cite{rubin2005causal} complements Pearl's SCM \cite{Richardson2013SingleWI} (cf. Box \ref{box1}) by providing a detailed list of necessary assumptions for a correct causal graph.

\item {\bf Selecting causal ML method.} The choice of causal ML method depends on the causal question, the available data, and the assumptions. For causal discovery questions, constraint methods (e.g., the PC and FCI algorithms) \cite{spirtes2000causation} and score methods (e.g., GES) \cite{chickering2002optimal} make minimal data assumptions but do not guarantee the discovery of the full causal graph. There exist variations for both cross-sectional \cite{glymour2019review} and time-series \cite{camps2023discovering,runge2019detecting} data. Asymmetry methods \cite{zhang2015estimation} make additional (process) assumptions regarding the types of functions that describe the cause-effect relationships to guarantee that they are all discovered. 

Answering causal effect questions involves estimating two key quantities: the Average Treatment Effect (ATE) and the Conditional Average Treatment Effect (CATE). The ATE summarizes the treatment's effectiveness by comparing average outcomes between treated and untreated groups, e.g., answering {\em `What is the average impact of organic farming (treatment) in Europe (entire population)?'}  
To estimate the ATE from observational data, researchers employ techniques such as matching \cite{rubin2006matched}, where similar treated and control groups are paired for comparison or propensity score methods \cite{rosenbaum1983central} which account for the probability of an individual receiving treatment. 
 
The CATE adopts a more granular approach by focusing on specific subgroups defined by observed covariates, e.g., answering {\em `How does the impact of organic farming (treatment) vary across countries in Europe (population subgroups)'} 
By estimating CATE, researchers can identify which subgroups benefit most from the treatment, allowing for the individualization of treatment recommendations. Techniques like the X-learner \cite{kunzel2019metalearners} and Double Machine Learning (DML) \cite{chernozhukov2018double} are commonly employed.

\item {\bf Checking robustness.} When we use observational data instead of data from randomized experiments to answer causal questions, we need to make assumptions about the data and the process that generated them. Robustness checks must be made to stress-test the plausibility of these assumptions. For causal discovery, this involves contrasting discovered relationships with domain knowledge. It may also involve testing the methods with artificially generated data, such as that generated with process-based models, and for which ground-truth causal graphs are available. Refutation tests \cite{sharma2021dowhy} are used for causal effect estimation to check if the assumptions made result in coherent estimates. Typical refutation methods involve manipulating the data so that the impact of this manipulation on results is predictable, provided the analysis is correct. For example,  noise variables can be added as additional confounder variables to check that the estimated effect remains stable. 
 
\end{itemize}

\subsection{Causality for improved ML predictions}\label{causality-aware}

Predictive ML models do not consider the causal structure of agricultural systems, thus they cannot adapt predictions accordingly when changes across space or time occur (cf. Box \ref{box1}). 
Causality-aware ML leverages the idea that associations coming from direct cause-effect relationships are the most stable (see apps in Sec. \ref{predict}). By selecting these direct causes, models can rely on relationships that persist across different environments or contexts \cite{cui2022stable}. This enhances the models' robustness to interventions, spatio-temporal generalizability, and explainability. This selection can be guided by prior knowledge of the causal graph or through data-driven techniques such as Invariant Causal Prediction (ICP) \cite{peters2016causal}, which aims to discover features that exhibit consistent predictive capacity under various experimental settings, environments, regimes, or interventions. However, solely relying on direct causes may be too conservative for some problems. Anchor Regression \cite{rothenhausler2021anchor} introduces a relaxation by incorporating weighted features that may extend beyond direct causes. This method allows the model to leverage additional information while prioritizing causal relationships, thus balancing accuracy and stability. Invariant Risk Minimization (IRM) \cite{arjovsky2019invariant} and Risk extrapolation (REx) \cite{krueger2021out} are other methods that leverage causal principles to enhance model stability. 

\section{Applications of causal ML in agriculture}\label{sec4}

Causal ML is about answering causal queries (like the ones in Box \ref{box2}), avoiding being right for the wrong reasons. 
We next showcase eight indicative applications of causal ML for sustainable agriculture (see Fig. \ref{fig2}); from advancing science to assisting policymakers, empowering farmers, and improving models. We characterize them with estimated levels of leverage, adoption risk, and impact. By leverage, we mean that the application can be realized and provide major changes and improvements with relatively small or focused efforts. Adoption risk refers to the challenges in getting scientists, policymakers, or farmers to use the outputs of the applications. Impact signifies the application's potential to create significant positive outcomes in the targeted areas.\\

\begin{mdframed}[backgroundcolor=gray!30]
\customlabel{box2}{2}
\textbf{Box 2: Questions that causal ML can answer.} 
\end{mdframed} 

\begin{mdframed}[backgroundcolor=gray!20]
\small Scientists \includegraphics[height=10pt]{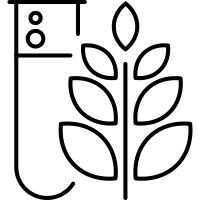}, policymakers \includegraphics[height=10pt]{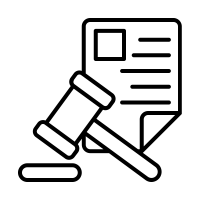}, farmers \includegraphics[height=10pt]{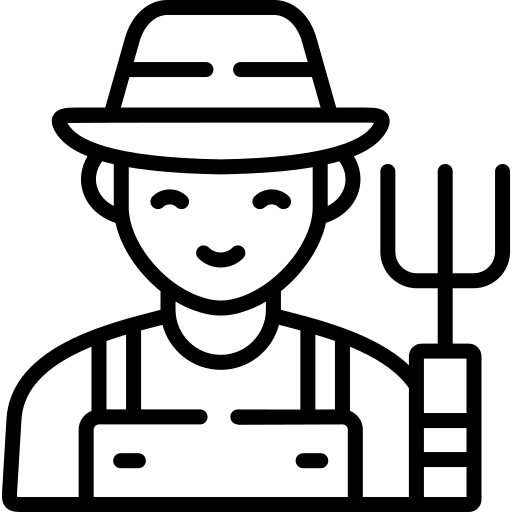}, and general public \includegraphics[height=10pt]{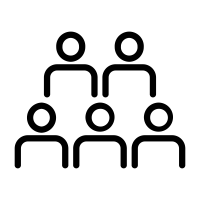}
\end{mdframed}

\begin{mdframed}[backgroundcolor=gray!10]
\noindent{\bf Causal discovery}
\begin{itemize}
    \item \textbf{What is the relationship between X and Y?} \includegraphics[height=10pt]{figures/icon1.png}
Does El Ni\~no Southern Oscillation (ENSO) cause soil moisture anomalies in Southern and Eastern Africa?
\item \textbf{What are the drivers of complex systems?} \includegraphics[height=10pt]{figures/icon1.png}, \includegraphics[height=10pt]{figures/icon2.png}
Which are the primary causal factors driving cropland expansion in the Amazon across varying socio-economic contexts? Are there contrasting mechanisms in the food security system for different districts in Somalia?
\end{itemize}
\noindent{\bf Causal effect}
\begin{itemize}
\item\textbf{What is the impact of climate change?} \includegraphics[height=10pt]{figures/icon1.png}, \includegraphics[height=10pt]{figures/icon2.png}
How do increased temperatures impact agricultural productivity? 
\item \textbf{What is the impact of human interventions? } \includegraphics[height=10pt]{figures/icon2.png}, \includegraphics[height=10pt]{figures/icon4.png}
What is the impact of humanitarian aid on food insecurity in Africa?
\item\textbf{What is the impact of extreme weather events?} \includegraphics[height=10pt]{figures/icon1.png}, \includegraphics[height=10pt]{figures/icon2.png}
Do dry spells drive crop failures in Europe?
\item\textbf{What is the impact of agricultural practices?} \includegraphics[height=10pt]{figures/icon2.png} 
Which regions benefit more from organic farming and what drives the differences in effects? 
How will the effect of practices be affected in the future under different climatic conditions? 

\item \textbf{Is a digital farm recommendation tool effective?}  \includegraphics[height=10pt]{figures/icon3.png} 

Will sowing on the recommended day increase my yield? What would be the benefit of investing in this digital tool?
\end{itemize}
\end{mdframed}

\subsection{Advancing science}\label{discovery}

\paragraph{Understanding complex systems}

\begin{figure}[t!]
  \centering
  \includegraphics[width=1 \textwidth]{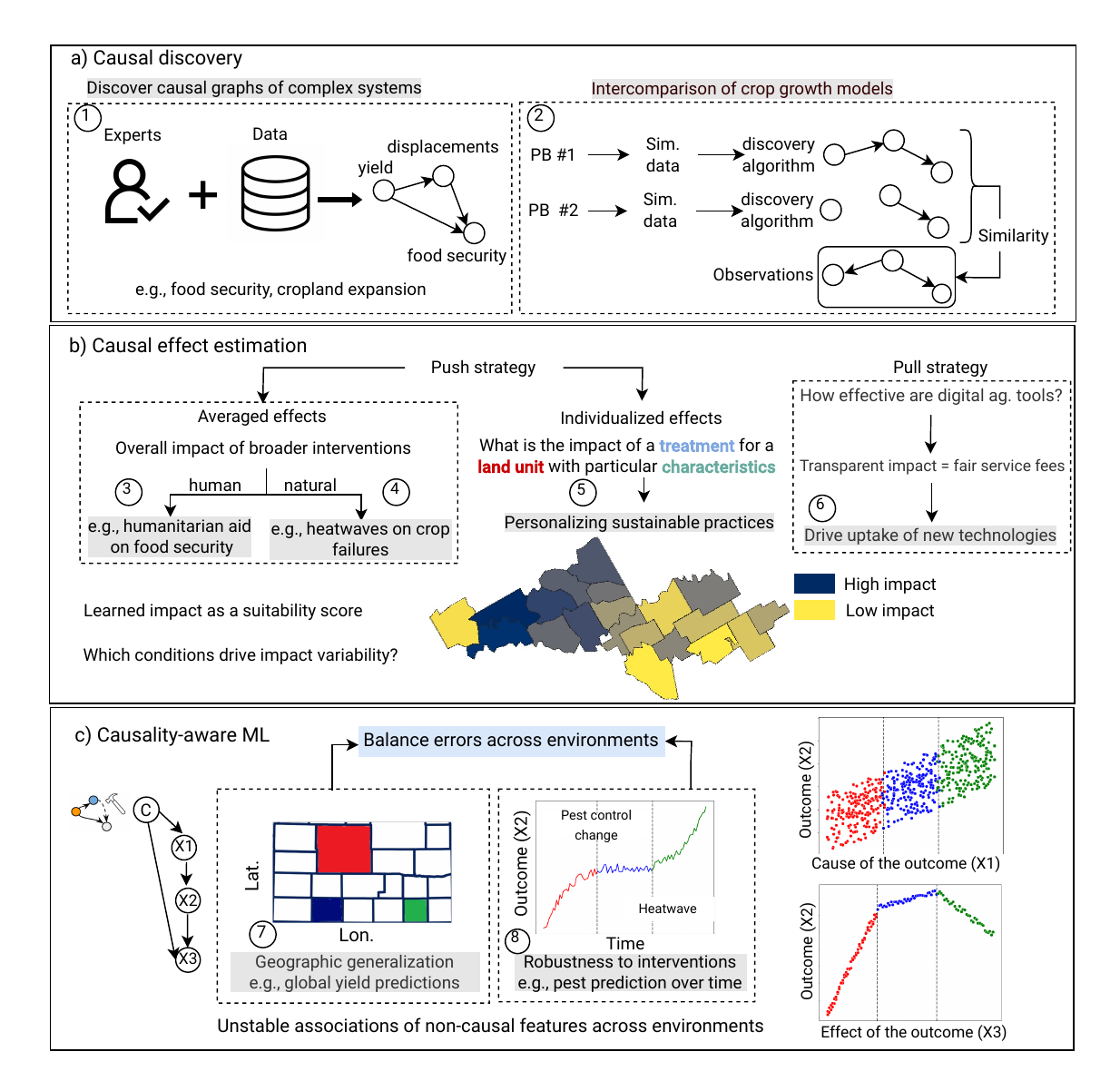}
\caption{\textbf{Applications of causal ML for agriculture.} Panel a) Causal discovery applications: Data-driven causal discovery (1) unveils causal mechanisms in complex systems like food security, enhancing domain expertise, and (2) evaluates process-based (PBs) models by comparing causal graphs from model simulations and observational data. 
Panel b) Causal effect estimation applications: Support evidence-based decisions by evaluating the impact of (3) human actions and (4) climate/weather events on sustainability outcomes. (5) Sustainable practices can be spatially tailored by estimating each land unit's individualized impact based on their characteristics. Which factors are responsible for the differences in impact? 
Panel c) Causality-aware ML applications: 
To achieve geographic generalization (7) and robustness to interventions (8) in predictive models, it is key to balance errors across various environments (e.g., geographic areas, anomalous weather events - variable C). This helps identify causal features, like X1, which maintain a stable relationship with the outcome, X2, under different conditions.}

  \label{fig2}
\end{figure}

\noindent Food security exemplifies complex systems with causal mechanisms that vary across scales, contexts, and time. Next to environmental stressors (e.g., persistent droughts), socio-economic factors such as poverty, poor market access, and lack of employment drive chronic food insecurity at the regional level \cite{misselhorn2005}. Local shocks, like climate extremes, and food price spikes drive sudden drops in regional food availability and exacerbate food security issues. Additionally, global market dynamics and trade routes, influenced by regional disruptions, impact food prices and availability elsewhere \cite{fyles2016}. The complexity of these systems challenges domain experts. Data-driven causal discovery techniques can complement expertise by analyzing large datasets to uncover hidden or emerging patterns (Fig. \ref{fig2}, app 1). For instance, researchers \cite{mwebaze2010causal} 
used a committee of causal discovery algorithms that collectively voted on the causal relationships between socioeconomic factors and famine risk in Uganda. Capturing the relationships in a causal graph enhances scientific understanding of underlying mechanisms and how they vary, aids in improving predictions of famine across different environments and contexts (cf. Sec. \ref{predict}), and facilitates the estimation of effects of various interventions on food security (cf. Sec. \ref{policy}). We assess this application as having low leverage, a medium level of adoption risk among domain scientists, and a moderate impact.

\paragraph{Intercomparison of crop growth models}

\noindent Crop growth models are essential for predicting yields, simulating environmental impacts, and optimizing resource use. However, current crop models exhibit large differences in their sensitivities to drivers such as carbon dioxide, temperature, and water \cite{muller_substantial_2024}, leading to uncertainty in projections of future global yields \cite{jagermeyr_climate_2021}. The Agricultural Model Intercomparison and Improvement Project (\href{AgMIP}{https://agmip.org/}) aims to improve crop growth models for assessing climate impacts on agriculture by conducting extensive protocol-based model intercomparisons. Inspired by the work of \cite{nowack2020causal}, who applied causal discovery algorithms to compare climate models, a similar approach could be implemented for the inter-comparison of crop growth models (Fig. \ref{fig2}, app 2). Simulations generated by various process-based crop growth models serve as inputs for causal discovery algorithms. The resulting causal graphs can be then compared using specific causal graph metrics with those derived from observational data to evaluate alignment with the actual causal structure of crop growth. This method offers an objective pathway for process-oriented model evaluation and improvement that goes beyond sensitivity analyses and enhances the reliability and accuracy of crop growth predictions. We assess this application as having high leverage, since crop growth models are already in place, a medium level of adoption risk among domain scientists, and a moderate impact.

\clearpage
\subsection{Assisting policymakers}\label{policy}

\paragraph{Impact of human and natural interventions}
 
\noindent Assessing the impact of policies, farmer decisions, or subsidies (treatments) on sustainability outcomes in agriculture is vital for evidence-based decision-making  (Fig. \ref{fig2}, app 3). For example, we can assess the impact of humanitarian aid, where cash transfers provide financial assistance to individuals and households during extreme droughts in the Horn of Africa \cite{cerdabautista2024assessing}. By estimating the ATE, we can measure the overall impact of these transfers on food security indicators, such as the Integrated Food Security Phase Classification (IPC), thus providing humanitarian organizations with evidence to design actionable responses. 
Shifting perspective, it is also useful to cast natural phenomena as the treatment of interest. For example, by understanding how extreme weather events, such as heatwaves or droughts, affect crop production, we can develop risk management strategies and support farmers in adapting (Fig. \ref{fig2}, app 4). Quantifying the overall impact of broader human and natural interventions is assessed as having high leverage, a medium level of adoption risk among policymakers, and a high impact. 

\paragraph{Personalizing sustainable practices}

\noindent Achieving sustainability goals in agriculture requires a shift from traditional one-size-fits-all approaches to geospatially tailored strategies \cite{giannarakis2022personalizing}. The impact of agricultural practices varies across different regions due to local conditions. Leveraging CATE methods facilitates the assessment of these varying effects (Fig. \ref{fig2}, app 5) which in turn enables the prioritization of the most beneficial practices for each land unit. For example, \cite{giannarakis2022towards} measured the CATE of crop rotation on net primary productivity (NPP) and proposed using it as a suitability score for applying crop rotation. They also explored which variables drove effect heterogeneity and showed that high min. temperatures combined with low water deficit increased the positive impact of crop rotation. We assess this application as having low leverage, as it requires large amounts of high-quality local data, a high level of adoption risk among policymakers and farmers, and a high impact. 

\subsection{Empowering farmers} \label{farmers}

\paragraph{Assessment of the effectiveness of digital agriculture} 

\noindent Digital agriculture tools, such as decision support systems and smart farming technologies, hold great promise for enhancing sustainability. However, low adoption rates remain a challenge. Evaluating their impact on farmers' profits can promote wider adoption (see Fig. \ref{fig2}, app 6). For instance, \cite{tsoumas2023evaluating} estimated a 12-17\% increase in crop yield for farmers who followed optimal day-of-sowing recommendations. Analyses like this can provide clear evidence, helping farmers make informed decisions and building trust in these technologies. Quantifying the effectiveness of digital agriculture tools also supports fair service fees and allows farmers to conduct accurate cost-benefit analyses. 
We assess this application as having low leverage due to data access issues, a medium level of adoption risk among farmers, and a moderate impact.

\subsection{Improving predictive modeling}\label{predict}

\noindent Causality-aware ML can improve predictive modeling in agriculture by addressing challenges related to model stability \cite{cui2022stable} (Fig. \ref{fig2}, app 7 and 8). For instance,  global information on national-level crop yield forecasting is key for food security crisis prevention and response planning. However, the uneven distribution of available training data, typically limited in time and quality for food-insecure countries, poses challenges for supervised models. Causality-aware ML methods (see Sec. \ref{causality-aware}), which focus on causal, stable variables, can offer a better geographic generalization for these models \cite{liu2022measure}. 
Additionally, these methods can offer robustness to interventions over time. For example, they can improve the resilience of pest prediction models when faced with changes in pest control strategies or anomalous environmental events \cite{tsoumas2023causality}. We assess these applications as having high leverage, as they capitalize on existing predictive models and their datasets, a low level of adoption risk among data modelers, and a moderate impact.

\section{Discussion and outlook}

We highlighted the potential of causal ML in promoting agricultural sustainability and addressing food security problems through two key approaches: enhancing the stability of predictive models (causality for ML) and answering causal questions (ML for causality). The applications we discussed demonstrate how causal ML can generate actionable insights to empower stakeholders, from farmers to global agricultural managers. 
We introduced a causal ML workflow to ensure the effective and reliable application of causal ML for sustainable agroecosystems.
Five elements should be worked out and integrated, all coming with their challenges and opportunities:

\noindent \textbf{Causal question.}  In pursuit of effective agricultural decisions, addressing the causal questions inherent in agroecosystem management is key. However, explicit causal language necessary for articulating these questions is largely absent in agriculture. Domain experts possess valuable knowledge that can be translated into a causal graph, which, when faithfully constructed, can be used alongside ML methods to answer causal queries. Correctly defining the causal question is important;  it must be relevant, precise, and feasible so that variables are measured accurately and appropriately \cite{pearl2009causality}.

\noindent \textbf{Data}. Agroecosystems are complex to model, predict, and understand, and involve nonlinearities and non-stationarities \cite{camps2023discovering,runge2023causal, runge2019inferring}. Effective analysis requires careful curation and harmonization of diverse data sources, including multi-modal Earth observation data and economic and social variables. High-resolution economic or social indicators like conflicts, GDP, and crop prices are often uncertain, less accessible, and inconsistently recorded compared to environmental data. The temporal and spatial scales of observation are important considerations. For example, in shorter time frames, weather can cause fluctuations in crop yield, whereas over longer periods, the impact of agriculture on local climate can be measured. Data limitations are a significant concern, especially because in causal graphs, the variables at the nodes are conceptual and may not be directly observable. It is important to minimize the gap between proxies and the original concepts to maintain the integrity of the analysis.

\noindent \textbf{Assumptions.} When randomization of treatment allocation is not possible, making specific assumptions is necessary to move from correlations to causal conclusions (cf. Box \ref{box1} and Sec. \ref{sec:2}). Structural, process, and statistical assumptions should be transparently laid out to ensure appropriate integration with data curation and method selection within the workflow. Assessing the plausibility of these assumptions is a necessary step \cite{sharma2021dowhy}; it relies heavily on expert domain knowledge. Their adequacy may also be evaluated by checking the performance of the entire causal ML workflow during validation. An important type of assumption not discussed in Sec. \ref{sec:2} regards how causal structure leads to spatial dependence. If spatial assumptions are not adequate this may lead to spatial biases \cite{akbari2023spatial}. 
For example, chemical pesticides used by one farmer may impact the water
quality for irrigation in neighboring fields as they seep into the groundwater (spillover).
    
\noindent \textbf{\bf Methods.} Once the data curation and assumption elements have been established to answer the causal question of interest, selecting a method involves choosing one that aligns with these elements. For example, for causal discovery questions if we are unwilling to assume there is no unmeasured confounding then we must choose a method such as FCI \cite{spirtes2000causation} that relaxes this assumption.  Gaps in the causal ML literature for agriculture include: scalable algorithms for mixed data types, high-dimensional data, and non-stationary causal relations \cite{runge2019inferring}; methods that account for unit interactions \cite{akbari2023spatial}; and causal representation learning methods for raw unstructured data \cite{scholkopf2022causality}.
    
\noindent \textbf{\bf Validation.} \textit{Causal discovery:} Establishing a database of causal graphs for agriculture based on expert knowledge is recommended. These graphs can then serve as a benchmark for evaluating different causal discovery algorithms and guide the definition of downstream causal effect estimation tasks. \textit{Effect estimation:} Ground truth data for effect estimates is typically not available. Instead, process-based models can be utilized: ground-truth causal effects can be extracted directly from their equations. Additionally, these equations can be used to generate artificial data. The causal effect estimation method is then evaluated on its ability to recover the ground-truth effects from the artificial data accurately. Another option is to combine observational effect estimates with randomized experiments \cite{kluger2022combining}. \textit{Causality-aware ML:} Model stability in causality-aware predictive models should be evaluated based on metrics like geographic generalization, transferability, or robustness to policy changes.
    
\noindent We recommend engagement in benchmark activities like those conducted by \href{AgML/AgMIP}{https://www.agml.org/} \cite{sweet2024} and making use of systematic evaluation platforms like \href{causeme}{https://causeme.uv.es} \cite{munoz2020causeme}. These initiatives foster collaboration, standardization, and continuous improvement in validating and evaluating (causal) ML methods in geosciences and agriculture.\\

\noindent Integrating causal thinking into agroecosystem sustainability is not only sensible but necessary, as it provides a robust framework for making informed decisions with significant environmental, economic, and societal implications.

\bibliography{sn-article}


\begin{thebibliography}{56}
\ifx \bisbn   \undefined \def \bisbn  #1{ISBN #1}\fi
\ifx \binits  \undefined \def \binits#1{#1}\fi
\ifx \bauthor  \undefined \def \bauthor#1{#1}\fi
\ifx \batitle  \undefined \def \batitle#1{#1}\fi
\ifx \bjtitle  \undefined \def \bjtitle#1{#1}\fi
\ifx \bvolume  \undefined \def \bvolume#1{\textbf{#1}}\fi
\ifx \byear  \undefined \def \byear#1{#1}\fi
\ifx \bissue  \undefined \def \bissue#1{#1}\fi
\ifx \bfpage  \undefined \def \bfpage#1{#1}\fi
\ifx \blpage  \undefined \def \blpage #1{#1}\fi
\ifx \burl  \undefined \def \burl#1{\textsf{#1}}\fi
\ifx \doiurl  \undefined \def \doiurl#1{\url{https://doi.org/#1}}\fi
\ifx \betal  \undefined \def \betal{\textit{et al.}}\fi
\ifx \binstitute  \undefined \def \binstitute#1{#1}\fi
\ifx \binstitutionaled  \undefined \def \binstitutionaled#1{#1}\fi
\ifx \bctitle  \undefined \def \bctitle#1{#1}\fi
\ifx \beditor  \undefined \def \beditor#1{#1}\fi
\ifx \bpublisher  \undefined \def \bpublisher#1{#1}\fi
\ifx \bbtitle  \undefined \def \bbtitle#1{#1}\fi
\ifx \bedition  \undefined \def \bedition#1{#1}\fi
\ifx \bseriesno  \undefined \def \bseriesno#1{#1}\fi
\ifx \blocation  \undefined \def \blocation#1{#1}\fi
\ifx \bsertitle  \undefined \def \bsertitle#1{#1}\fi
\ifx \bsnm \undefined \def \bsnm#1{#1}\fi
\ifx \bsuffix \undefined \def \bsuffix#1{#1}\fi
\ifx \bparticle \undefined \def \bparticle#1{#1}\fi
\ifx \barticle \undefined \def \barticle#1{#1}\fi
\bibcommenthead
\ifx \bconfdate \undefined \def \bconfdate #1{#1}\fi
\ifx \botherref \undefined \def \botherref #1{#1}\fi
\ifx \url \undefined \def \url#1{\textsf{#1}}\fi
\ifx \bchapter \undefined \def \bchapter#1{#1}\fi
\ifx \bbook \undefined \def \bbook#1{#1}\fi
\ifx \bcomment \undefined \def \bcomment#1{#1}\fi
\ifx \oauthor \undefined \def \oauthor#1{#1}\fi
\ifx \citeauthoryear \undefined \def \citeauthoryear#1{#1}\fi
\ifx \endbibitem  \undefined \def \endbibitem {}\fi
\ifx \bconflocation  \undefined \def \bconflocation#1{#1}\fi
\ifx \arxivurl  \undefined \def \arxivurl#1{\textsf{#1}}\fi
\csname PreBibitemsHook\endcsname

\bibitem{francis2003agroecology}
\begin{barticle}
\bauthor{\bsnm{Francis}, \binits{C.}},
\bauthor{\bsnm{Lieblein}, \binits{G.}},
\bauthor{\bsnm{Gliessman}, \binits{S.}},
\bauthor{\bsnm{Breland}, \binits{T.}},
\bauthor{\bsnm{Creamer}, \binits{N.}},
\bauthor{\bsnm{Harwood}, \binits{R.}},
\bauthor{\bsnm{Salomonsson}, \binits{L.}},
\bauthor{\bsnm{Helenius}, \binits{J.}},
\bauthor{\bsnm{Rickerl}, \binits{D.}},
\bauthor{\bsnm{Salvador}, \binits{R.}},
\bauthor{\bsnm{Wiedenhoeft}, \binits{M.}},
\bauthor{\bsnm{Simmons}, \binits{S.}},
\bauthor{\bsnm{Allen}, \binits{P.}},
\bauthor{\bsnm{Altieri}, \binits{M.}},
\bauthor{\bsnm{Flora}, \binits{C.}},
\bauthor{\bsnm{Poincelot}, \binits{R.}}:
\batitle{Agroecology: The ecology of food systems}.
\bjtitle{Journal of Sustainable Agriculture}
\bvolume{22},
\bfpage{99}--\blpage{118}
(\byear{2003}).
\doiurl{10.1300/J064v22n03_10}
\end{barticle}
\endbibitem

\bibitem{pretty2008agricultural}
\begin{barticle}
\bauthor{\bsnm{Pretty}, \binits{J.}}:
\batitle{Agricultural sustainability: concepts, principles and evidence}.
\bjtitle{Philosophical Transactions of the Royal Society of London, Series B}
\bvolume{363},
\bfpage{447}--\blpage{465}
(\byear{2013})
\end{barticle}
\endbibitem

\bibitem{godfray2010food}
\begin{barticle}
\bauthor{\bsnm{Godfray}, \binits{H.C.J.}},
\bauthor{\bsnm{Beddington}, \binits{J.R.}},
\bauthor{\bsnm{Crute}, \binits{I.R.}}:
\batitle{Food security: the challenge of feeding 9 billion people}.
\bjtitle{Science}
\bvolume{327}(\bissue{5967}),
\bfpage{812}--\blpage{818}
(\byear{2010})
\end{barticle}
\endbibitem

\bibitem{van2021meta}
\begin{barticle}
\bauthor{\bsnm{Van~Dijk}, \binits{M.}},
\bauthor{\bsnm{Morley}, \binits{T.}},
\bauthor{\bsnm{Rau}, \binits{M.L.}},
\bauthor{\bsnm{Saghai}, \binits{Y.}}:
\batitle{A meta-analysis of projected global food demand and population at risk of hunger for the period 2010--2050}.
\bjtitle{Nature Food}
\bvolume{2}(\bissue{7}),
\bfpage{494}--\blpage{501}
(\byear{2021})
\end{barticle}
\endbibitem

\bibitem{eswaran2019land}
\begin{botherref}
\oauthor{\bsnm{Eswaran}, \binits{H.}},
\oauthor{\bsnm{Lal}, \binits{R.}},
\oauthor{\bsnm{Reich}, \binits{P.}}:
Land degradation: an overview.
Response to land degradation,
20--35
(2019)
\end{botherref}
\endbibitem

\bibitem{ivanovich2023future}
\begin{barticle}
\bauthor{\bsnm{Ivanovich}, \binits{C.C.}},
\bauthor{\bsnm{Sun}, \binits{T.}},
\bauthor{\bsnm{Gordon}, \binits{D.R.}},
\bauthor{\bsnm{Ocko}, \binits{I.B.}}:
\batitle{Future warming from global food consumption}.
\bjtitle{Nature Climate Change}
\bvolume{13}(\bissue{3}),
\bfpage{297}--\blpage{302}
(\byear{2023})
\end{barticle}
\endbibitem

\bibitem{takahashi2020technology}
\begin{botherref}
\oauthor{\bsnm{Takahashi}, \binits{K.}},
\oauthor{\bsnm{Muraoka}, \binits{R.}},
\oauthor{\bsnm{Otsuka}, \binits{K.}}:
Technology adoption, impact, and extension in developing countries’ agriculture: A review of the recent literature.
Agricultural Economics
\textbf{51}
(2019).
\doiurl{10.1111/agec.12539}
\end{botherref}
\endbibitem

\bibitem{peer2020european}
\begin{botherref}
\oauthor{\bsnm{Pe’er}, \binits{G.}},
\oauthor{\bparticle{van~der} \bsnm{Werf}, \binits{W.}},
\oauthor{\bsnm{Piro-Smith}, \binits{E.}}:
The eu’s common agricultural policy could be spent much more efficiently to address challenges
(2020)
\end{botherref}
\endbibitem

\bibitem{altieri2020agroecology}
\begin{barticle}
\bauthor{\bsnm{Altieri}, \binits{M.A.}},
\bauthor{\bsnm{Nicholls}, \binits{C.I.}}:
\batitle{Agroecology and the emergence of a post covid-19 agriculture}.
\bjtitle{Agriculture and Human Values}
\bvolume{37},
\bfpage{525}--\blpage{526}
(\byear{2020})
\end{barticle}
\endbibitem

\bibitem{umesha2018sustainable}
\begin{bchapter}
\bauthor{\bsnm{Umesha}, \binits{S.}},
\bauthor{\bsnm{Manukumar}, \binits{H.M.}},
\bauthor{\bsnm{Chandrasekhar}, \binits{B.}}:
\bctitle{Sustainable agriculture and food security}.
In: \bbtitle{Biotechnology for Sustainable Agriculture},
pp. \bfpage{67}--\blpage{92}.
\bpublisher{Elsevier},
\blocation{NL}
(\byear{2018})
\end{bchapter}
\endbibitem

\bibitem{gliessman2016transforming}
\begin{barticle}
\bauthor{\bsnm{Gliessman}, \binits{S.R.}}, \betal:
\batitle{Transforming food systems with agroecology}.
\bjtitle{Agroecology and Sustainable Food Systems}
\bvolume{40}(\bissue{3}),
\bfpage{187}--\blpage{189}
(\byear{2016})
\end{barticle}
\endbibitem

\bibitem{meshram2021machine}
\begin{barticle}
\bauthor{\bsnm{Meshram}, \binits{V.}},
\bauthor{\bsnm{Patil}, \binits{K.}},
\bauthor{\bsnm{Meshram}, \binits{V.}},
\bauthor{\bsnm{Hanchate}, \binits{D.}},
\bauthor{\bsnm{Ramkteke}, \binits{S.}}:
\batitle{Machine learning in agriculture domain: A state-of-art survey}.
\bjtitle{Artificial Intelligence in the Life Sciences}
\bvolume{1},
\bfpage{100010}
(\byear{2021})
\end{barticle}
\endbibitem

\bibitem{scholkopf2022causality}
\begin{bchapter}
\bauthor{\bsnm{Sch{\"o}lkopf}, \binits{B.}}:
\bctitle{Causality for machine learning}.
In: \bbtitle{Probabilistic and Causal Inference: The Works of Judea Pearl},
pp. \bfpage{765}--\blpage{804}
(\byear{2022})
\end{bchapter}
\endbibitem

\bibitem{kaddour2022causal}
\begin{botherref}
\oauthor{\bsnm{Kaddour}, \binits{J.}},
\oauthor{\bsnm{Preux}, \binits{P.}}:
Causal machine learning: A survey and open problems.
arXiv preprint arXiv:2206.15475
(2022)
\end{botherref}
\endbibitem

\bibitem{runge2023modern}
\begin{barticle}
\bauthor{\bsnm{Runge}, \binits{J.}}:
\batitle{Modern causal inference approaches to investigate biodiversity-ecosystem functioning relationships}.
\bjtitle{Nature Communications}
\bvolume{14}(\bissue{1}),
\bfpage{1917}
(\byear{2023})
\end{barticle}
\endbibitem

\bibitem{fougere2019causal}
\begin{botherref}
\oauthor{\bsnm{Fougère}, \binits{D.}},
\oauthor{\bsnm{Jacquemet}, \binits{N.}}:
Causal inference and impact evaluation.
Economie et Statistique/Economics and Statistics
(2019)
\end{botherref}
\endbibitem

\bibitem{runge2023causal}
\begin{barticle}
\bauthor{\bsnm{Runge}, \binits{J.}},
\bauthor{\bsnm{Gerhardus}, \binits{A.}},
\bauthor{\bsnm{Varando}, \binits{G.}},
\bauthor{\bsnm{Eyring}, \binits{V.}},
\bauthor{\bsnm{Camps-Valls}, \binits{G.}}:
\batitle{Causal inference for time series}.
\bjtitle{Nature Reviews Earth \& Environment}
\bvolume{4}(\bissue{7}),
\bfpage{487}--\blpage{505}
(\byear{2023})
\end{barticle}
\endbibitem

\bibitem{pearl2009causality}
\begin{bbook}
\bauthor{\bsnm{Pearl}, \binits{J.}}:
\bbtitle{Causality}.
\bpublisher{Cambridge university press},
\blocation{Cambridge, UK and New York, NY, USA}
(\byear{2009})
\end{bbook}
\endbibitem

\bibitem{petersen1994agricultural}
\begin{bbook}
\bauthor{\bsnm{Petersen}, \binits{R.G.}}:
\bbtitle{Agricultural Field Experiments: Design and Analysis}.
\bpublisher{CRC Press},
\blocation{FL, USA}
(\byear{1994})
\end{bbook}
\endbibitem

\bibitem{kluger2022combining}
\begin{barticle}
\bauthor{\bsnm{Kluger}, \binits{D.M.}},
\bauthor{\bsnm{Owen}, \binits{A.B.}},
\bauthor{\bsnm{Lobell}, \binits{D.B.}}:
\batitle{Combining randomized field experiments with observational satellite data to assess the benefits of crop rotations on yields}.
\bjtitle{Environmental Research Letters}
\bvolume{17}(\bissue{4}),
\bfpage{044066}
(\byear{2022})
\end{barticle}
\endbibitem

\bibitem{rothwell2005external}
\begin{barticle}
\bauthor{\bsnm{Rothwell}, \binits{P.M.}}:
\batitle{External validity of randomised controlled trials:“to whom do the results of this trial apply?”}.
\bjtitle{The Lancet}
\bvolume{365}(\bissue{9453}),
\bfpage{82}--\blpage{93}
(\byear{2005})
\end{barticle}
\endbibitem

\bibitem{li2022deep}
\begin{barticle}
\bauthor{\bsnm{Li}, \binits{J.}},
\bauthor{\bsnm{Hong}, \binits{D.}},
\bauthor{\bsnm{Gao}, \binits{L.}},
\bauthor{\bsnm{Yao}, \binits{J.}},
\bauthor{\bsnm{Zheng}, \binits{K.}},
\bauthor{\bsnm{Zhang}, \binits{B.}},
\bauthor{\bsnm{Chanussot}, \binits{J.}}:
\batitle{Deep learning in multimodal remote sensing data fusion: A comprehensive review}.
\bjtitle{International Journal of Applied Earth Observation and Geoinformation}
\bvolume{112},
\bfpage{102926}
(\byear{2022})
\end{barticle}
\endbibitem

\bibitem{rubin2005causal}
\begin{barticle}
\bauthor{\bsnm{Rubin}, \binits{D.B.}}:
\batitle{Causal inference using potential outcomes: Design, modeling, decisions}.
\bjtitle{Journal of the American Statistical Association}
\bvolume{100}(\bissue{469}),
\bfpage{322}--\blpage{331}
(\byear{2005})
\end{barticle}
\endbibitem

\bibitem{Richardson2013SingleWI}
\begin{bchapter}
\bauthor{\bsnm{Richardson}, \binits{T.S.}}:
\bctitle{{Single World Intervention Graphs (SWIGs): A Unification of the Counterfactual and Graphical Approaches to Causality}}.
(\byear{2013})
\end{bchapter}
\endbibitem

\bibitem{spirtes2000causation}
\begin{botherref}
\oauthor{\bsnm{Spirtes}, \binits{P.}},
\oauthor{\bsnm{Glymour}, \binits{C.}},
\oauthor{\bsnm{Scheines}, \binits{R.}}:
Causation, prediction, and search
(2000)
\end{botherref}
\endbibitem

\bibitem{chickering2002optimal}
\begin{barticle}
\bauthor{\bsnm{Chickering}, \binits{D.M.}},
\bauthor{\bsnm{Heckerman}, \binits{D.}}:
\batitle{Optimal structure identification with greedy search}.
\bjtitle{Journal of machine learning research}
\bvolume{3}(\bissue{Dec}),
\bfpage{507}--\blpage{554}
(\byear{2002})
\end{barticle}
\endbibitem

\bibitem{glymour2019review}
\begin{barticle}
\bauthor{\bsnm{Glymour}, \binits{C.}},
\bauthor{\bsnm{Zhang}, \binits{K.}},
\bauthor{\bsnm{Spirtes}, \binits{P.}}:
\batitle{Review of causal discovery methods based on graphical models}.
\bjtitle{Frontiers in genetics}
\bvolume{10},
\bfpage{524}
(\byear{2019})
\end{barticle}
\endbibitem

\bibitem{camps2023discovering}
\begin{barticle}
\bauthor{\bsnm{Camps-Valls}, \binits{G.}},
\bauthor{\bsnm{Gerhardus}, \binits{A.}},
\bauthor{\bsnm{Ninad}, \binits{U.}},
\bauthor{\bsnm{Varando}, \binits{G.}},
\bauthor{\bsnm{Martius}, \binits{G.}},
\bauthor{\bsnm{Balaguer-Ballester}, \binits{E.}},
\bauthor{\bsnm{Vinuesa}, \binits{R.}},
\bauthor{\bsnm{Diaz}, \binits{E.}},
\bauthor{\bsnm{Zanna}, \binits{L.}},
\bauthor{\bsnm{Runge}, \binits{J.}}:
\batitle{Discovering causal relations and equations from data}.
\bjtitle{Physics Reports}
\bvolume{1044}(\bissue{1}),
\bfpage{1}--\blpage{68}
(\byear{2023})
\end{barticle}
\endbibitem

\bibitem{runge2019detecting}
\begin{barticle}
\bauthor{\bsnm{Runge}, \binits{J.}},
\bauthor{\bsnm{Nowack}, \binits{P.}},
\bauthor{\bsnm{Kretschmer}, \binits{M.}},
\bauthor{\bsnm{Flaxman}, \binits{S.}},
\bauthor{\bsnm{Sejdinovic}, \binits{D.}}:
\batitle{Detecting and quantifying causal associations in large nonlinear time series datasets}.
\bjtitle{Science advances}
\bvolume{5}(\bissue{11}),
\bfpage{4996}
(\byear{2019})
\end{barticle}
\endbibitem

\bibitem{zhang2015estimation}
\begin{barticle}
\bauthor{\bsnm{Zhang}, \binits{K.}},
\bauthor{\bsnm{Wang}, \binits{Z.}},
\bauthor{\bsnm{Zhang}, \binits{J.}},
\bauthor{\bsnm{Sch{\"o}lkopf}, \binits{B.}}:
\batitle{On estimation of functional causal models: general results and application to the post-nonlinear causal model}.
\bjtitle{ACM Transactions on Intelligent Systems and Technology (TIST)}
\bvolume{7}(\bissue{2}),
\bfpage{1}--\blpage{22}
(\byear{2015})
\end{barticle}
\endbibitem

\bibitem{rubin2006matched}
\begin{barticle}
\bauthor{\bsnm{Rubin}, \binits{D.B.}}:
\batitle{Matched sampling for causal effects}.
\bjtitle{Journal of Computerized Tomography}
\bvolume{30}(\bissue{3}),
\bfpage{16}--\blpage{19}
(\byear{2006})
\end{barticle}
\endbibitem

\bibitem{rosenbaum1983central}
\begin{barticle}
\bauthor{\bsnm{Rosenbaum}, \binits{P.R.}},
\bauthor{\bsnm{Rubin}, \binits{D.B.}}:
\batitle{The central role of the propensity score in observational studies for causal effects}.
\bjtitle{Biometrika}
\bvolume{70}(\bissue{1}),
\bfpage{41}--\blpage{55}
(\byear{1983})
\end{barticle}
\endbibitem

\bibitem{kunzel2019metalearners}
\begin{barticle}
\bauthor{\bsnm{K{\"u}nzel}, \binits{S.}},
\bauthor{\bsnm{Sekhon}, \binits{J.S.}},
\bauthor{\bsnm{Bickel}, \binits{J.}},
\bauthor{\bsnm{Yu}, \binits{B.}},
\bauthor{\bsnm{Bennett}, \binits{C.}},
\bauthor{\bsnm{Xie}, \binits{M.}},
\bauthor{\bsnm{Tibshirani}, \binits{R.}}:
\batitle{Metalearners for estimating heterogeneous treatment effects using machine learning}.
\bjtitle{Proceedings of the National Academy of Sciences}
\bvolume{116}(\bissue{10}),
\bfpage{4156}--\blpage{4165}
(\byear{2019})
\end{barticle}
\endbibitem

\bibitem{chernozhukov2018double}
\begin{botherref}
\oauthor{\bsnm{Chernozhukov}, \binits{V.}},
\oauthor{\bsnm{Chetverikov}, \binits{D.}},
\oauthor{\bsnm{Demirer}, \binits{M.}},
\oauthor{\bsnm{Duflo}, \binits{E.}},
\oauthor{\bsnm{Hansen}, \binits{C.}},
\oauthor{\bsnm{Newey}, \binits{W.}},
\oauthor{\bsnm{Robins}, \binits{J.}}:
Double/debiased machine learning for treatment and structural parameters.
Oxford University Press Oxford, UK
(2018)
\end{botherref}
\endbibitem

\bibitem{sharma2021dowhy}
\begin{botherref}
\oauthor{\bsnm{Sharma}, \binits{A.}},
\oauthor{\bsnm{Syrgkanis}, \binits{V.}},
\oauthor{\bsnm{Zhang}, \binits{C.}},
\oauthor{\bsnm{K{\i}c{\i}man}, \binits{E.}}:
Dowhy: Addressing challenges in expressing and validating causal assumptions.
arXiv preprint arXiv:2108.13518
(2021)
\end{botherref}
\endbibitem

\bibitem{cui2022stable}
\begin{barticle}
\bauthor{\bsnm{Cui}, \binits{P.}},
\bauthor{\bsnm{Athey}, \binits{S.}}:
\batitle{Stable learning establishes some common ground between causal inference and machine learning}.
\bjtitle{Nature Machine Intelligence}
\bvolume{4}(\bissue{2}),
\bfpage{110}--\blpage{115}
(\byear{2022})
\end{barticle}
\endbibitem

\bibitem{peters2016causal}
\begin{barticle}
\bauthor{\bsnm{Peters}, \binits{J.}},
\bauthor{\bsnm{B{\"u}hlmann}, \binits{P.}},
\bauthor{\bsnm{Meinshausen}, \binits{N.}}:
\batitle{Causal inference by using invariant prediction: identification and confidence intervals}.
\bjtitle{Journal of the Royal Statistical Society Series B: Statistical Methodology}
\bvolume{78}(\bissue{5}),
\bfpage{947}--\blpage{1012}
(\byear{2016})
\end{barticle}
\endbibitem

\bibitem{rothenhausler2021anchor}
\begin{barticle}
\bauthor{\bsnm{Rothenh{\"a}usler}, \binits{D.}},
\bauthor{\bsnm{Meinshausen}, \binits{N.}},
\bauthor{\bsnm{B{\"u}hlmann}, \binits{P.}},
\bauthor{\bsnm{Peters}, \binits{J.}}:
\batitle{Anchor regression: Heterogeneous data meet causality}.
\bjtitle{Journal of the Royal Statistical Society Series B: Statistical Methodology}
\bvolume{83}(\bissue{2}),
\bfpage{215}--\blpage{246}
(\byear{2021})
\end{barticle}
\endbibitem

\bibitem{arjovsky2019invariant}
\begin{botherref}
\oauthor{\bsnm{Arjovsky}, \binits{M.}},
\oauthor{\bsnm{Bottou}, \binits{L.}},
\oauthor{\bsnm{Gulrajani}, \binits{I.}},
\oauthor{\bsnm{Lopez-Paz}, \binits{D.}}:
Invariant risk minimization.
arXiv preprint arXiv:1907.02893
(2019)
\end{botherref}
\endbibitem

\bibitem{krueger2021out}
\begin{bchapter}
\bauthor{\bsnm{Krueger}, \binits{D.}},
\bauthor{\bsnm{Caballero}, \binits{E.}},
\bauthor{\bsnm{Jacobsen}, \binits{J.-H.}},
\bauthor{\bsnm{Zhang}, \binits{A.}},
\bauthor{\bsnm{Binas}, \binits{J.}},
\bauthor{\bsnm{Zhang}, \binits{D.}},
\bauthor{\bsnm{Le~Priol}, \binits{R.}},
\bauthor{\bsnm{Courville}, \binits{A.}}:
\bctitle{Out-of-distribution generalization via risk extrapolation (rex)}.
In: \bbtitle{International Conference on Machine Learning},
pp. \bfpage{5815}--\blpage{5826}
(\byear{2021}).
\bcomment{PMLR}
\end{bchapter}
\endbibitem

\bibitem{misselhorn2005}
\begin{barticle}
\bauthor{\bsnm{Misselhorn}, \binits{A.A.}}:
\batitle{What drives food insecurity in southern africa? a meta-analysis of household economy studies}.
\bjtitle{Global Environmental Change}
\bvolume{15}(\bissue{1}),
\bfpage{33}--\blpage{43}
(\byear{2005}).
\doiurl{10.1016/j.gloenvcha.2004.11.003}
\end{barticle}
\endbibitem

\bibitem{fyles2016}
\begin{bchapter}
\bauthor{\bsnm{Fyles}, \binits{H.}},
\bauthor{\bsnm{Madramootoo}, \binits{C.}}:
\bctitle{{Key Drivers of Food Insecurity}}.
In: \beditor{\bsnm{Madramootoo}, \binits{C.}} (ed.)
\bbtitle{Emerging Technologies for Promoting Food Security}.
\bsertitle{Woodhead Publishing Series in Food Science, Technology and Nutrition},
pp. \bfpage{1}--\blpage{19}.
\bpublisher{Woodhead Publishing},
\blocation{Oxford}
(\byear{2016}).
\doiurl{10.1016/B978-1-78242-335-5.00001-9}.
\burl{https://www.sciencedirect.com/science/article/pii/B9781782423355000019}
\end{bchapter}
\endbibitem

\bibitem{mwebaze2010causal}
\begin{bchapter}
\bauthor{\bsnm{Mwebaze}, \binits{E.}},
\bauthor{\bsnm{Okori}, \binits{W.}},
\bauthor{\bsnm{Quinn}, \binits{J.A.}}:
\bctitle{Causal structure learning for famine prediction}.
In: \bbtitle{2010 AAAI Spring Symposium Series}
(\byear{2010})
\end{bchapter}
\endbibitem

\bibitem{muller_substantial_2024}
\begin{barticle}
\bauthor{\bsnm{Müller}, \binits{C.}},
\bauthor{\bsnm{Jägermeyr}, \binits{J.}},
\bauthor{\bsnm{Franke}, \binits{J.A.}},
\bauthor{\bsnm{Ruane}, \binits{A.C.}},
\bauthor{\bsnm{Balkovic}, \binits{J.}},
\bauthor{\bsnm{Ciais}, \binits{P.}},
\bauthor{\bsnm{Dury}, \binits{M.}},
\bauthor{\bsnm{Falloon}, \binits{P.}},
\bauthor{\bsnm{Folberth}, \binits{C.}},
\bauthor{\bsnm{Hank}, \binits{T.}},
\bauthor{\bsnm{Hoffmann}, \binits{M.}},
\bauthor{\bsnm{Izaurralde}, \binits{R.C.}},
\bauthor{\bsnm{Jacquemin}, \binits{I.}},
\bauthor{\bsnm{Khabarov}, \binits{N.}},
\bauthor{\bsnm{Liu}, \binits{W.}},
\bauthor{\bsnm{Olin}, \binits{S.}},
\bauthor{\bsnm{Pugh}, \binits{T.A.M.}},
\bauthor{\bsnm{Wang}, \binits{X.}},
\bauthor{\bsnm{Williams}, \binits{K.}},
\bauthor{\bsnm{Zabel}, \binits{F.}},
\bauthor{\bsnm{Elliott}, \binits{J.W.}}:
\batitle{Substantial {Differences} in {Crop} {Yield} {Sensitivities} {Between} {Models} {Call} for {Functionality}-{Based} {Model} {Evaluation}}.
\bjtitle{Earth's Future}
\bvolume{12}(\bissue{3}),
\bfpage{2023}--\blpage{003773}
(\byear{2024}).
\doiurl{10.1029/2023EF003773}
\end{barticle}
\endbibitem

\bibitem{jagermeyr_climate_2021}
\begin{barticle}
\bauthor{\bsnm{Jägermeyr}, \binits{J.}},
\bauthor{\bsnm{Müller}, \binits{C.}},
\bauthor{\bsnm{Ruane}, \binits{A.C.}},
\bauthor{\bsnm{Elliott}, \binits{J.}},
\bauthor{\bsnm{Balkovic}, \binits{J.}},
\bauthor{\bsnm{Castillo}, \binits{O.}},
\bauthor{\bsnm{Faye}, \binits{B.}},
\bauthor{\bsnm{Foster}, \binits{I.}},
\bauthor{\bsnm{Folberth}, \binits{C.}},
\bauthor{\bsnm{Franke}, \binits{J.A.}},
\bauthor{\bsnm{Fuchs}, \binits{K.}},
\bauthor{\bsnm{Guarin}, \binits{J.R.}},
\bauthor{\bsnm{Heinke}, \binits{J.}},
\bauthor{\bsnm{Hoogenboom}, \binits{G.}},
\bauthor{\bsnm{Iizumi}, \binits{T.}},
\bauthor{\bsnm{Jain}, \binits{A.K.}},
\bauthor{\bsnm{Kelly}, \binits{D.}},
\bauthor{\bsnm{Khabarov}, \binits{N.}},
\bauthor{\bsnm{Lange}, \binits{S.}},
\bauthor{\bsnm{Lin}, \binits{T.-S.}},
\bauthor{\bsnm{Liu}, \binits{W.}},
\bauthor{\bsnm{Mialyk}, \binits{O.}},
\bauthor{\bsnm{Minoli}, \binits{S.}},
\bauthor{\bsnm{Moyer}, \binits{E.J.}},
\bauthor{\bsnm{Okada}, \binits{M.}},
\bauthor{\bsnm{Phillips}, \binits{M.}},
\bauthor{\bsnm{Porter}, \binits{C.}},
\bauthor{\bsnm{Rabin}, \binits{S.S.}},
\bauthor{\bsnm{Scheer}, \binits{C.}},
\bauthor{\bsnm{Schneider}, \binits{J.M.}},
\bauthor{\bsnm{Schyns}, \binits{J.F.}},
\bauthor{\bsnm{Skalsky}, \binits{R.}},
\bauthor{\bsnm{Smerald}, \binits{A.}},
\bauthor{\bsnm{Stella}, \binits{T.}},
\bauthor{\bsnm{Stephens}, \binits{H.}},
\bauthor{\bsnm{Webber}, \binits{H.}},
\bauthor{\bsnm{Zabel}, \binits{F.}},
\bauthor{\bsnm{Rosenzweig}, \binits{C.}}:
\batitle{Climate impacts on global agriculture emerge earlier in new generation of climate and crop models}.
\bjtitle{Nature Food}
\bvolume{2}(\bissue{11}),
\bfpage{873}--\blpage{885}
(\byear{2021}).
\doiurl{10.1038/s43016-021-00400-y}
\end{barticle}
\endbibitem

\bibitem{nowack2020causal}
\begin{barticle}
\bauthor{\bsnm{Nowack}, \binits{P.}},
\bauthor{\bsnm{Kretschmer}, \binits{M.}},
\bauthor{\bsnm{Marotzke}, \binits{J.}}:
\batitle{Causal networks for climate model evaluation and constrained projections}.
\bjtitle{Nature Communications}
\bvolume{11}(\bissue{1}),
\bfpage{1415}
(\byear{2020})
\end{barticle}
\endbibitem

\bibitem{cerdabautista2024assessing}
\begin{botherref}
\oauthor{\bsnm{Cerd\`a-Bautista}, \binits{J.}},
\oauthor{\bsnm{T\'arraga}, \binits{J.M.}},
\oauthor{\bsnm{Sitokonstantinou}, \binits{V.}},
\oauthor{\bsnm{Camps-Valls}, \binits{G.}}:
Assessing the Causal Impact of Humanitarian Aid on Food Security
(2024)
\end{botherref}
\endbibitem

\bibitem{giannarakis2022personalizing}
\begin{bchapter}
\bauthor{\bsnm{Giannarakis}, \binits{G.}},
\bauthor{\bsnm{Sitokonstantinou}, \binits{V.}},
\bauthor{\bsnm{Lorilla}, \binits{R.S.}},
\bauthor{\bsnm{Kontoes}, \binits{C.}}:
\bctitle{Personalizing sustainable agriculture with causal machine learning}.
In: \bbtitle{NeurIPS 2022 Workshop on Tackling Climate Change with Machine Learning}
(\byear{2022})
\end{bchapter}
\endbibitem

\bibitem{giannarakis2022towards}
\begin{bchapter}
\bauthor{\bsnm{Giannarakis}, \binits{G.}},
\bauthor{\bsnm{Sitokonstantinou}, \binits{V.}},
\bauthor{\bsnm{Lorilla}, \binits{R.S.}},
\bauthor{\bsnm{Kontoes}, \binits{C.}}:
\bctitle{Towards assessing agricultural land suitability with causal machine learning}.
In: \bbtitle{Proceedings of the IEEE/CVF Conference on Computer Vision and Pattern Recognition},
pp. \bfpage{1442}--\blpage{1452}
(\byear{2022})
\end{bchapter}
\endbibitem

\bibitem{tsoumas2023evaluating}
\begin{bchapter}
\bauthor{\bsnm{Tsoumas}, \binits{I.}},
\bauthor{\bsnm{Giannarakis}, \binits{G.}},
\bauthor{\bsnm{Sitokonstantinou}, \binits{V.}},
\bauthor{\bsnm{Koukos}, \binits{A.}},
\bauthor{\bsnm{Loka}, \binits{D.}},
\bauthor{\bsnm{Bartsotas}, \binits{N.}},
\bauthor{\bsnm{Athanasiadis}, \binits{I.}}:
\bctitle{Evaluating digital agriculture recommendations with causal inference}.
In: \bbtitle{Proceedings of the AAAI Conference on Artificial Intelligence},
vol. \bseriesno{37},
pp. \bfpage{14514}--\blpage{14522}
(\byear{2023})
\end{bchapter}
\endbibitem

\bibitem{liu2022measure}
\begin{bchapter}
\bauthor{\bsnm{Liu}, \binits{J.}},
\bauthor{\bsnm{Wu}, \binits{J.}},
\bauthor{\bsnm{Pi}, \binits{R.}},
\bauthor{\bsnm{Xu}, \binits{R.}},
\bauthor{\bsnm{Zhang}, \binits{X.}},
\bauthor{\bsnm{Li}, \binits{B.}},
\bauthor{\bsnm{Cui}, \binits{P.}}:
\bctitle{Measure the predictive heterogeneity}.
In: \bbtitle{The Eleventh International Conference on Learning Representations}
(\byear{2022})
\end{bchapter}
\endbibitem

\bibitem{tsoumas2023causality}
\begin{bchapter}
\bauthor{\bsnm{Tsoumas}, \binits{I.}},
\bauthor{\bsnm{Sitokonstantinou}, \binits{V.}},
\bauthor{\bsnm{Giannarakis}, \binits{G.}},
\bauthor{\bsnm{Lampiri}, \binits{E.}},
\bauthor{\bsnm{Athanassiou}, \binits{C.}},
\bauthor{\bsnm{Camps-Valls}, \binits{G.}},
\bauthor{\bsnm{Athanasiadis}, \binits{I.}}:
\bctitle{Causality and explainability for trustworthy integrated pest management}.
In: \bbtitle{NeurIPS 2023 Workshop on Tackling Climate Change with Machine Learning}
(\byear{2023})
\end{bchapter}
\endbibitem

\bibitem{runge2019inferring}
\begin{barticle}
\bauthor{\bsnm{Runge}, \binits{J.}},
\bauthor{\bsnm{Bathiany}, \binits{S.}},
\bauthor{\bsnm{Bollt}, \binits{E.}},
\bauthor{\bsnm{Camps-Valls}, \binits{G.}},
\bauthor{\bsnm{Coumou}, \binits{D.}},
\bauthor{\bsnm{Deyle}, \binits{E.}},
\bauthor{\bsnm{Glymour}, \binits{C.}},
\bauthor{\bsnm{Kretschmer}, \binits{M.}},
\bauthor{\bsnm{Mahecha}, \binits{M.D.}},
\bauthor{\bsnm{Mu{\~n}oz-Mar{\'\i}}, \binits{J.}}, \betal:
\batitle{Inferring causation from time series in earth system sciences}.
\bjtitle{Nature communications}
\bvolume{10}(\bissue{1}),
\bfpage{2553}
(\byear{2019})
\end{barticle}
\endbibitem

\bibitem{akbari2023spatial}
\begin{barticle}
\bauthor{\bsnm{Akbari}, \binits{K.}},
\bauthor{\bsnm{Winter}, \binits{S.}},
\bauthor{\bsnm{Tomko}, \binits{M.}}:
\batitle{Spatial causality: A systematic review on spatial causal inference}.
\bjtitle{Geographical Analysis}
\bvolume{55}(\bissue{1}),
\bfpage{56}--\blpage{89}
(\byear{2023})
\end{barticle}
\endbibitem

\bibitem{sweet2024}
\begin{botherref}
\oauthor{\bsnm{Sweet}, \binits{L.}},
\oauthor{\bsnm{Athanasiadis}, \binits{I.N.}},
\oauthor{\bparticle{van} \bsnm{Bree}, \binits{R.}},
\oauthor{\bsnm{Castellano}, \binits{A.}},
\oauthor{\bsnm{Martre}, \binits{P.}},
\oauthor{\bsnm{Paudel}, \binits{D.}},
\oauthor{\bsnm{Ruane}, \binits{A.C.}},
\oauthor{\bsnm{Zscheischler}, \binits{J.}}:
{ Transdisciplinary coordination is essential for advancing agricultural modeling with machine learning}.
Submitted manuscript
(2024)
\end{botherref}
\endbibitem

\bibitem{munoz2020causeme}
\begin{botherref}
\oauthor{\bsnm{Munoz-Mar{\'\i}}, \binits{J.}},
\oauthor{\bsnm{Mateo}, \binits{G.}},
\oauthor{\bsnm{Runge}, \binits{J.}},
\oauthor{\bsnm{Camps-Valls}, \binits{G.}}:
Causeme: An online system for benchmarking causal discovery methods.
In Preparation
(2020)
\end{botherref}
\endbibitem

\end{thebibliography}
\section*{Competing Interests}
The authors declare no competing interests.

\clearpage
\scriptsize



\end{document}